# Evaluation of GPT and BERT-based models on identifying protein-protein interactions in biomedical text

Hasin Rehana[1,*], Nur Bengisu Çam[2,*], Mert Basmaci[2], Jie Zheng[3], Christianah Jemiyo[4], Yongqun He[3,5], Arzucan Özgür[2,§], and Junguk Hur[4,§]

[1]Computer Science Graduate Program, University of North Dakota, Grand Forks, North Dakota, 58202, USA
[2]Department of Computer Engineering, Bogazici University, 34342 Istanbul, Turkey
[3]Unit for Laboratory Animal Medicine, Department of Microbiology and Immunology, University of Michigan, Ann Arbor, Michigan, 48109, USA
[4]Department of Biomedical Sciences, University of North Dakota School of Medicine and Health Sciences, Grand Forks, North Dakota, 58202, USA
[5]Center for Computational Medicine and Bioinformatics, University of Michigan, Ann Arbor, Michigan, 48109, USA

*These authors contributed equally to this work.
§ Corresponding authors
    Junguk Hur, PhD
    University of North Dakota Department of Biomedical Sciences
    1301 N Columbia Rd. Stop 9037. Rm 130W
    Grand Forks, ND 58202
    Phone: 701-777-6814
    Email: junguk.hur@med.und.edu

    Arzucan Özgür, PhD
    Department of Computer Engineering
    Bogazici University
    34342 Bebek, Istanbul, Turkey
    Email: arzucan.ozgur@boun.edu.tr

# Abstract

Detecting protein-protein interactions (PPIs) is crucial for understanding genetic mechanisms, disease pathogenesis, and drug design. However, with the fast-paced growth of biomedical literature, there is a growing need for automated and accurate extraction of PPIs to facilitate scientific knowledge discovery. Pre-trained language models, such as generative pre-trained transformers (GPT) and bidirectional encoder representations from transformers (BERT), have shown promising results in natural language processing (NLP) tasks. We evaluated the performance of PPI identification of multiple GPT and BERT models using three manually curated gold-standard corpora: Learning Language in Logic (LLL) with 164 PPIs in 77 sentences, Human Protein Reference Database with 163 PPIs in 145 sentences, and Interaction Extraction Performance Assessment with 335 PPIs in 486 sentences. BERT-based models achieved the best overall performance, with BioBERT achieving the highest recall (91.95%) and F1-score (86.84%) and PubMedBERT achieving the highest precision (85.25%). Interestingly, despite not being explicitly trained for biomedical texts, GPT-4 achieved commendable performance, comparable to the top-performing BERT models. It achieved a precision of 88.37%, a recall of 85.14%, and an F1-score of 86.49% on the LLL dataset. These results suggest that GPT models can effectively detect PPIs from text data, offering promising avenues for application in biomedical literature mining. Further research could explore how these models might be fine-tuned for even more specialized tasks within the biomedical domain.

# Introduction

Protein-protein interactions (PPIs) are essential for numerous biological functions, especially DNA replication and transcription, signal pathways, cell metabolism, and converting genotype to phenotype. Gaining insight into these interactions enhances understanding of the biological processes, pathways, and networks underlying healthy and diseased states. Various public PPI databases exist [1-3], including the PPI data collected from low-to-mid throughput experiments such as yeast-two-hybrid or immunoprecipitation pull-down or high-throughput screening assays. However, these resources are incomplete and do not cover all potential PPIs. Due to the rapid growth of scientific literature, manual extraction of PPIs has become increasingly challenging, which necessitates automated text mining approaches that do not require human participation.

Natural language processing (NLP) is a focal area in computer science, increasingly applied in various domains, including biomedical research, which has experienced massive growth in recent years. Relation extraction, a widely used NLP method, aims to identify relationships between two or more entities in biomedical text, supporting the automatic analysis of documents in this domain [4]. Advances in deep learning (DL), such as convolutional neural networks (CNNs) [5, 6] and recurrent neural networks (RNNs) [7, 8], as well as in NLP, have enabled success in biomedical text mining to discover interactions between protein entities. Building on these advancements, a comprehensive study reviewed recent research from 2021 to 2023, specifically focusing on applying deep learning methodologies in predicting PPIs. This review highlighted the emergence of Graph Neural Networks (GNNs) as a powerful tool for modeling complex protein interaction networks alongside the established roles of CNNs and Autoencoders. These methodologies have been instrumental in deciphering complex patterns and interactions within biological datasets, offering new insights into the dynamics of protein interactions and the broader mechanisms governing biological systems.

Besides biomedical text datasets, several studies [9, 10] have experimented with protein sequences. For instance, one study [9] utilized ensemble CNN architectures with residual connections on protein sequence datasets to mitigate the gradient vanishing/exploding problem that can happen while training deep models. The research introduced a novel CNN method for predicting PPI, enhancing positive samples' accuracy by incorporating residue binding propensity, thereby addressing the cost and time challenges of experimental methods [11]. Meanwhile, Soleymani et al. devised a framework combining an autoencoder with a deep CNN, which achieved significant accuracy and efficiency in PPI prediction [12]. Concurrently, another study [10] used ensemble deep neural networks (DNNs) to protein sequence datasets. The field has also seen advancements through pretraining large neural language models, leading to substantial improvements in various NLP problems [9]. Following the seminar work "Attention Is All You Need" [10], transformer architectures have set new benchmarks in various NLP tasks, including relation extraction in the biomedical domain [13].

After the development of transformer architecture [10], transformer-based models like bidirectional encoder representation transformer (BERT) [14], a type of masked language model, emerged. These models, known as large language models (LLMs), focus on understanding language and semantics. LLMs are pre-trained on vast amounts of data and can be fine-tuned for various tasks. Recent studies suggest that LLMs excel at context zero-shot and few-shot learning [15], analyzing, producing, and comprehending human languages. LLMs' massive data processing capabilities can be employed to identify connections and trends among textual elements. Some of the recent studies [10, 16-18], used domain-specific pre-trained BERT models on the five PPI gold-standard datasets: Learning Language in Logic (LLL)

[19], a subset of Human Protein Reference Database (HPRD50) [20], and Interaction Extraction Performance Assessment (IEPA) [21], AIMED [22], and Bio Information Extraction Resource (BioInfer) [23]. Park et al. [16] used transformer-based architectures to capture protein entity relationships in five PPI gold-standard datasets. Roy et al. [17] improved classification performance using tree transformers and a graph neural network. Warikoo et al. [18] created a Lexically aware Transformer-based Bidirectional Encoder Representation model.

Another type of LLM is autoregressive language models, including generative pre-trained transformer (GPT), an advanced AI language model that generates human-like text by acquiring linguistic patterns and structures. GPT [24] is a series of language models developed by OpenAI in 2018 based on transformer architecture [14]. The transformer model consists of an encoder that generates concealed representations and a decoder that produces output sequences using multi-head attention, which prioritizes data over inductive biases, facilitating large-scale pretraining and parallelization. The self-attention mechanism enables neural networks to determine the importance of input elements, making it ideal for language translation, text classification, and text generation. The first version of GPT, GPT-1, had 117 million parameters. It was trained using a large corpus of text data, including Wikipedia (https://en.wikipedia.org/), Common Crawl (https://commoncrawl.org/the-data/), and OpenWebText (https://skylion007.github.io/OpenWebTextCorpus/). GPT-2 [25] significantly improved over its predecessor with roughly ten times bigger 1.5 billion parameters. It was trained on a larger corpus of text data, including web pages and books, and can generate more coherent and convincing language responses. GPT-3 [26] was trained with 175 billion parameters, including an enormous corpus of text data, web pages, books, and academic articles. GPT-3 has demonstrated outstanding performance in a wide range of NLP tasks, such as language translation, chatbot development, and content generation. On November 30[th], 2022, OpenAI released ChatGPT, a natural and engaging conversation tool capable of producing contextually relevant responses based on text data. ChatGPT was fine-tuned on the GPT-3.5 series. On March 14th, 2023, OpenAI introduced its most advanced and cutting-edge system to date, GPT-4, which has surpassed its predecessors by producing more dependable outcomes. GPT-4 [27] is capable of producing, modifying, and cooperating with users in various creative and technical writing tasks, such as songwriting, screenplay creation, and imitating user writing styles [28]. The advancements in GPT models, from GPT-3 to GPT-4, showcase the rapid progress in NLP, opening up a wide range of applications [29, 30].

Several studies [31, 32] have been published evaluating the performance of GPT models for problem-solving on various standardized tests. It has been shown that they can achieve performance comparable to or even better than humans [33] and can pass high-level professional standardized tests such as the Bar test [34], the Chinese Medical Practitioners examination [35], and the Japanese National Nurse examination [36]. Another study [37] evaluates that ChatGPT achieves the equivalent of a passing score for a third-year medical student in the United States Medical Licensing Examination. A group of researchers explored the performance of advanced LLMs, including ChatGPT, GPT-4, and Google Bard, in mastering complex neurosurgery examination material and found that GPT-4 achieved a score of 82.6%, outperforming ChatGPT and Google Bard. A recent study assessed the efficacy of GPT in relation to specific standardized admissions tests in the UK. These findings revealed that GPT demonstrates greater proficiency in tasks requiring language comprehension and processing yet exhibits limitations in applications involving scientific and mathematical knowledge [38]. To the best of our knowledge, no study has evaluated the effectiveness of GPT models in extracting PPIs from biomedical texts. In this article, we present a thorough evaluation of the PPI identification performance of the GPT-based models and compare these with the state-of-the-art BERT-based NLP models.

# Methods

## Language Models

We evaluated two autoregressive language models (GPT-3.5 and GPT-4), each with six variations, and three masked language models (BioBERT, PubMedBERT, and SciBERT).

## Autoregressive Language Models

The GPT architecture includes layers with self-attention mechanisms, fully connected layers, and layer normalization, reducing computational time and preventing overfitting during training [23]. **Figure 1** illustrates a brief history of GPT models introduced by OpenAI in recent years. In our study, we have used the updated version released on June 13th, 2023, for GPT-3.5 and GPT-4 (gpt-3.5-0613 and gpt-4-0613, respectively). The architecture and number of parameters for GPT models are summarized in **Supplementary Table 1**, including the GPT-3.5 and GPT-4 models, which were included in the current study.

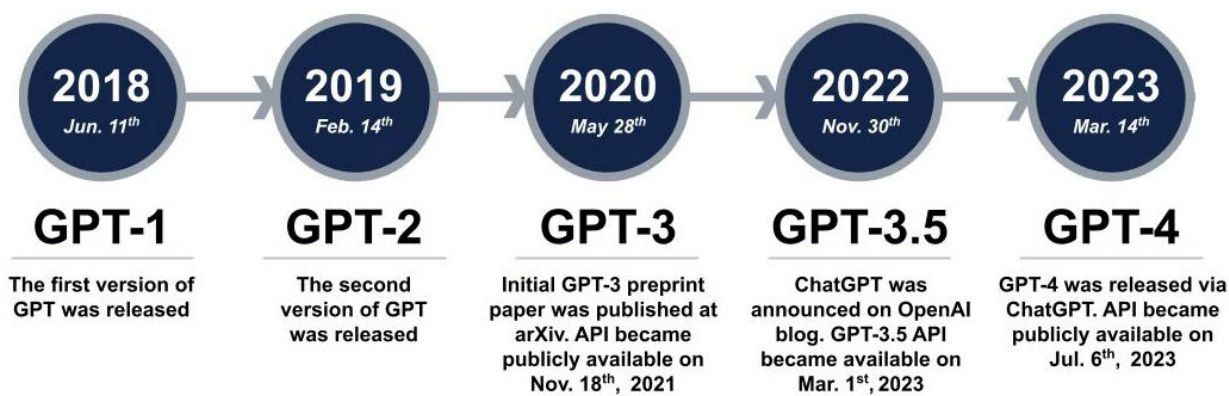

**Figure 1. Evolution of GPT Models**. (GPT: generative pre-trained transformer. API: application programming interface.)

## Masked Language Models

Three different BERT-based models were included in the current study (**Table 1**). Even though there are many different BERT-based models that are either trained or fine-tuned on biomedical corpora, we have only selected the BioBERT [15], SciBERT [39], and PubMedBERT [13] models. It is simply because these are among the most commonly used BERT-based models for biomedical text processing.

- BioBERT [15]: a BERT model pre-trained on PubMed abstracts and PubMed Central (PMC) full-text articles for different NLP tasks to measure performance. The initial version, BioBERT v1.0, used >200K abstracts and >270K PMC articles. An expanded version of BioBERT v1.1 was fine-tuned using > 1M PubMed abstracts and was included in the current study. The model was accessed from the following HuggingFace repository: dmis-lab/biobert-v1.1 .
- SciBERT [39]: a BERT model pre-trained on random Semantic Scholar articles [40]. While pretraining with the articles, the entire text was used. The researchers created SCIVOCAB from scientific articles of the same size as BASEVOCAB, the BERT-base model's vocabulary. Uncased SCIVOCAB was used in the current study. The model was accessed from the following HuggingFace repository: allenai/scibert_scivocab_uncased .
- PubMedBERT [13]: a BERT model trained explicitly on the BLURB (Biomedical Language Understanding & Reasoning Benchmark). In this study, we used the PubMedBERT which was trained only on the abstracts. The model was accessed from the following HuggingFace repository: microsoft/BiomedNLP-PubMedBERT-base-uncased-abstract.

**Table 1: Specifications of BERT-based models**

| Model Name | Year Released | Architecture | Number of Parameters |
|---|---|---|---|
| BioBERT | 2019 | Encoder architecture of transformer with 12 layers and hidden size of 768 | 108 million |
| SciBERT | 2019 | Encoder architecture of transformer with 12 layers and hidden size of 768 | 108 million |
| PubMedBERT | 2020 | Encoder architecture of transformer with 12 layers and hidden size of 768 | 108 million |

## Dataset

In our study, we employed three widely used datasets for PPI extraction: LLL [19], IEPA [21], and HPRD50 [20]. Each of these gold-standard datasets offers a unique perspective and challenge in the field of biomedical NLP research, particularly in the area of PPI extraction. Originally released as part of the LLL shared task challenge in 2005, the LLL dataset is focused on extracting protein/gene interactions from biological abstracts related to *Bacillus subtilis*. The dataset contains 77 sentences with 164 manually annotated positive protein pairs, also referred to as positive pairs in this study. The IEPA dataset comprises nearly 300 abstracts from MEDLINE, utilizing specific biochemical noun queries. It includes 486 sentences with 335 positive pairs and 482 negative pairs. The dataset is a valuable resource for assessing the performance of PPI extraction models, given its diverse range of abstracts and detailed annotation of protein interactions. The HPRD50 dataset, a subset of the Human Protein Reference Database (HPRD), consists of 50 randomly selected abstracts. The dataset is annotated for various interactions, including direct physical interactions, regulatory relations, and modifications such as phosphorylation. It contains 145 sentences, with 163 positive pairs and 270 negative pairs. The dataset details are listed in **Table 2.**

**Table 2:** Number of positive and negative pairs in the sentences of the three datasets.

| Dataset | Number of Sentences | Positive Pairs | Negative Pairs | Total Pairs | Ratio of Positive and Negative PPI pairs |
|---|---|---|---|---|---|
| LLL | 77 | 164 | 166 | 330 | 1:1.0 |
| IEPA | 486 | 335 | 482 | 817 | 1:1.4 |
| HPRD50 | 145 | 163 | 270 | 433 | 1:1.6 |

We preprocessed the sentences in these gold-standard datasets. For the GPT models, protein dictionaries for each dataset were compiled by listing all the protein mentions and removing any duplicate instances from the list. We also applied preprocessing steps to ensure capturing all entities by removing punctuation marks, digit-only strings, and blank spaces and converting all the letters into lowercase, resulting in a normalized protein dictionary. Both the original and normalized versions of dictionaries are utilized as auxiliary resources in our experiments with GPT-based models. For the BERT-based models, similar to the prior work[41], we replaced the entity pair names with the *PROTEIN1* and *PROTEIN2* keywords. Other than the pair, the entity names in the sentence were substituted with the '*PROTEIN'* keyword. Although GPT models are not masked language models, we have used both the original sentences and PROTEIN-masked sentences for evaluating GPT-based models.

We extracted the necessary data from the gold-standard datasets and divided each into ten folds using a document-level folding strategy, as previously described [42]. The same ten folds were utilized for both GPT and BERT-based models. In the PROTEIN masked settings, sentences were repeated with the placeholders PROTEIN1, PROTEIN2, and PROTEIN positioned in various locations to represent different protein pairs within the same sentence. This procedure, however, resulted in minor variations of the same sentence, posing a potential challenge for GPT model interpretation. To mitigate this issue, we also implemented an N-fold partitioning strategy in our dataset processing. This approach was designed to ensure that no individual partition contained duplicated sentences with varied placements of protein placeholders, thereby reducing the likelihood of model confusion in distinguishing between these slight sentence variations.

## Prompt Engineering for GPT models

To extract PPIs from input sentences, we leveraged OpenAI's application programming interface (API) access for the GPT models. We carefully designed the prompts to generate well-structured and stable interactions with minimal post-processing steps. We initiated this process by developing a foundational prompt consisting of seven sections (**Table 3**).

**Table 3 Structured Breakdown of a Complete Prompt for step-by-step prompt Engineering.**

| Section number | Purpose | Details |
|---|---|---|
| 1 | Primary instruction about extracting PPI | Extract every pair of Protein-Protein Interaction (PPI) from the sentences provided as input, analyzing each sentence individually. |
| 2 | Proteins=Genes | For this task, 'Proteins' and 'Genes' are synonymous. |
| 3 | Handling multiple PPI pairs | If a sentence contains multiple PPI pairs, list each pair on a distinct row. |
| 4 | Output format | Please, format your results in CSV (comma-separated values) format with the following four columns: 'Sentence ID', 'Protein 1', 'Protein 2', and 'Interaction Type'. Ensure that no columns are left blank. |
| 5 | Output format column specification | Output Column Specifications:<br>'Sentence ID': The unique identifier for each sentence.<br>'Protein 1' and 'Protein 2': The entities in the sentence, representing the proteins or genes.<br>'Interaction Type': The type of interaction identified between the protein entities (e.g., 'binds to', 'inhibits'). |
| 6 | End-of-process indication | If all sentences have been processed successfully, the last row should only contain the word 'Done'. |
| 7 | Input sentences | Each input line contains a 'Sentence ID' and corresponding 'Sentence' that is needed to be analyzed for finding PPI.<br>Here are the sentences that you need to process: |

In our approach to prompt engineering, we emphasized creating a diverse array of prompt variations for each section to pinpoint the most effective formats. The core of this approach was in Section 1, which is critical as it contains the primary instructions for the PPI identification task. We developed a comprehensive set of 128 variations for this section. For the subsequent sections, we generated fewer variations: 54 for Section 2, 96 for Section 3, 72 each for Sections 4 and 5, 24 for Section 6, and 12 for Section 7. We incorporated the variations in the foundational prompt while maintaining the other sections constant during section-wise testing. Then, we performed a section-wise evaluation to find the best variation of prompt from each section. The F1 score was used as the performance metric to select an optimal prompt for further analysis. The evaluation process was sequential, examining each section individually to select the most effective prompt variation before progressing to the next. This systematic approach ensured that each section's best variation was carried forward into subsequent tests. After identifying the top prompt from each section, we conducted a secondary analysis of these seven prompts. This step was crucial to mitigate any potential temporal biases that might have arisen from varying times of API access. The detailed algorithm for this prompt engineering process is depicted in **Figure 2.** Given the extensive nature of prompt engineering and the associated costs, we opted to utilize GPT-3.5 for prompt engineering instead of the more expensive GPT-4.

**Figure 2: Algorithm for prompt engineering**

```
Algorithm 1 Optimization of Base Prompt
 1: currentPrompt ← INITIALFOUNDATIONALPROMPT
 2: bestPrompts ← empty list
 3: for each section S_i and number of variations N_i in
    [(S1, 128), (S2, 54), (S3, 96), (S4, 72), (S5, 72), (S6, 24), (S7, 12)] do
 4:     bestVariation ← null
 5:     highestF1Score ← −∞
 6:     for j ← 1 to N_i do
 7:         testPrompt ← GENERATEVARIATION(currentPrompt, S_i, j)
 8:         output ← APPLYGPT3.5(testPrompt)
 9:         f1Score ← EVALUATEF1SCORE(output)
10:         if f1Score > highestF1Score then
11:             highestF1Score ← f1Score
12:             bestVariation ← testPrompt
13:         end if
14:     end for
15:     Add bestVariation to bestPrompts
16:     currentPrompt ← bestVariation
17: end for
18: finalBasePrompt ← null
19: highestFinalF1Score ← −∞
20: for each prompt in bestPrompts do
21:     output ← APPLYGPT3.5(testPrompt)
22:     f1Score ← EVALUATEF1SCORE(output)
23:     if f1Score > highestFinalF1Score then
24:         highestFinalF1Score ← f1Score
25:         finalBasePrompt ← prompt
26:     end if
27: end for
28: return finalBasePrompt
```

To broaden the scope of our investigation, we introduced two additional prompt variations, incorporating a dictionary of the original or normalized Protein mentioned, respectively. This strategy was chosen to assess the impact of additional domain-specific information on GPT models, as they were not explicitly trained for biomedical analysis or PPI identification. The details of the protein dictionaries can be found in **Table 4.**

**Table 4: Details of Original and Normalized Dictionaries for LLL, IEPA and HPRD50 datasets**

|  | *Dictionary Type* | Dataset | | |
| --- | --- | --- | --- | --- |
|  |  | LLL | IEPA | HPRD50 |
| *Number of unique protein names* | Original | 122 | 130 | 189 |
|  | Normalized | 103 | 100 | 182 |
| *Average number of characters in protein names* | Original | 4.92 | 11.8 | 8.75 |
|  | Normalized | 4.58 | 10.46 | 7.93 |
| *Maximum number of characters in protein names* | Original | 9 | 54 | 48 |
|  | Normalized | 8 | 50 | 44 |
| *Minimum number of characters in protein names* | Original | 3 | 1 | 3 |
|  | Normalized | 3 | 1 | 3 |

For protein-masked inputs, we adapted the base prompt to suit the altered input-output pattern. This adaptation involved experimenting with variations in the different sections of the prompts, ultimately selecting the variant with the highest F1 score for our analysis. In all scenarios, while using GPT models, we provided the sentence IDs and sentences as input, along with a query. Due to the context window limitations of GPT models, we split some of the larger folds into two (HPRD50) or three (IEPA) parts when the total number of tokens in input and estimated output exceeds the maximum token limit of the models. However, we combined the output parts from the same folds together before calculating performance scores. For all the model settings, we conducted ten independent runs and calculated their averages to mitigate any biases. To avoid the potential bias of accessing the GPT models at different times, we have implemented parallel processing. This involved generating a randomized list of all possible combinations of dataset folds, models, and prompt types for each run, thereby ensuring a balanced and unbiased experimental setup.

## Temperature parameter optimization

OpenAI's API allows the modulation of the 'temperature' parameter in GPTs, which determines how greedy or creative the generative model is. The parameter ranges between 0 (the most precise) and 2 (the most creative) for GPT-3.5 and GPT-4. In our initial consideration for exploring the impact of temperature on PPI identification using the OpenAI API, we set the range of temperature settings from a minimum of 0 to a maximum of 2. However, we found that as the temperature increased above 1, the failure rate in generating output in the required format also increased along with performance degradation. Consequently, we revised our approach and limited the range to 0 to 1 for further analysis, taking into account the longer processing times associated with higher temperature settings.

## Performance evaluation

For BERT-based models, we fine-tuned these models in a 10-fold cross-validation setting, where the folds were created at the document level, as introduced above. This strategy employed document-level fold splitting, which ensured the sentences from one document were used only either in the training or testing set to avoid overfitting [43]. All the fine-tuning experiments were conducted on the Tesla A100. We experimented with 1e-5, 2e-5, and 5e-5 as a learning rate, 0.1, 0.01, and 0.001 as a weight decay, and 4, 6, 8, and 32 as a batch size. In order to provide fixed length input to the models, we used 128 and 256 as the maximum length to pad. When each example in each dataset was tokenized, the token length was more than 128. When the sentence's token length is more than 128, and the maximum length is 128, the sentence is truncated. Since we want to provide the whole sentence into the model, we used 256 as a maximum padding length, which resulted in better performance. Also, note that fine-tuning with a maximum padding length of 128 is faster compared to 256. We selected the best hyperparameters based on the average F1 score for models on all datasets' test splits. The best hyperparameters are shown in **Table 5**.

**Table 5: Hyperparameters used for K-Fold Cross Validation on BERT models.**

| Hyperparameter | Value |
| --- | --- |
| Optimizer | AdamW |
| Weight Decay | 0.01 |
| Learning Rate | 2e-5 |
| Batch Size | 4 |
| Epochs per Fold | 8 |
| Max Sentence Length | 256 |

# Results

## Interacting with GPT API

**Figure 3** illustrates a Python code segment to access GPT API and its output. The predicted interaction pairs were returned with corresponding Sentence IDs.

```
def call_GPT_API(query, Sentences, max_tokens, temperature):
    model = "gpt-3.5-turbo-0613"
    prompt = f"""
    {query}
    {Sentences}
    """
    messages = [{"role": "user", "content": prompt}]
    response = openai.ChatCompletion.create(
        model=model,
        messages=messages,
        max_tokens=max_tokens,
        temperature=temperature
    )
    output= response['choices'][0]['message']['content']
    return output
                    (A)
```

```
Sentence ID,Protein 1,Protein 2,Interaction Type
LLL.d8.s0,KatX,EsigmaF,depends on
LLL.d8.s2,sigmaB,sigmaF,overlap
LLL.d8.s3,katX,sigmaB,dependent
LLL.d10.s0,rocG,SigL,transcribed by
LLL.d10.s0,rocG,RocR,requires for its expression
LLL.d11.s0,phrC,sigmaH,controlling transcription
LLL.d30.s0,sigma(X),sigma(A),recognize
LLL.d38.s0,sigB,gsiB,member of
Done
                    (B)
```

**Figure 3. GPT API code and output.** (A) Python code segment for accessing OpenAI API. (B) an example output of GPT-4 for Fold 1.

## Prompt Engineering for GPT Models

The performance of all the variations of Prompts from each section is visualized in **Supplementary Figures 3 to 9.** The most effective prompt from each of the seven sections, identified through extensive testing, is presented in **Supplementary Table 2**. Among these 7 best prompts from each section, prompt 60 from Section 3 (P60_S3) performed best in terms of F1 score. So, we selected P60_S3 as the base prompt and adjusted that to get the other settings as mentioned in the Methods part. The final prompts of six different types of settings (base prompt, base prompt with protein dictionary, base prompt with normalized protein dictionary, PROTEIN masked prompt, PROTEIN masked prompt with no repeated sentence in same prompt, and Protein masked prompt with one input sentence at a time) are detailed in

**Supplementary Table 4**. However, the protein dictionaries mentioned in the tables under "with protein dictionary" and "with normalized protein dictionary" apply to the LLL dataset. For the other datasets, the specified dictionary is adapted accordingly.

## Temperature parameter optimization

A temperature of 0.0 demonstrated the highest overall performance of GPT-3.5-0613 with the most consistent output format. **Figure 4** depicts the score variation over a temperature range from 0 to 1.0. For GPT-4-0613, we have only explored the final base prompt due to the high expense of GPT-4. From **Figure 5**, it is visible that the precision, recall, and F1 scores are highest for temperature 0.0, although the differences are minor. However, the output's structure becomes slightly inconsistent at higher temperatures, necessitating additional post-processing. Therefore, in our current study, a temperature setting of 0.0 was chosen for both GPT-3.5 and GPT-4.

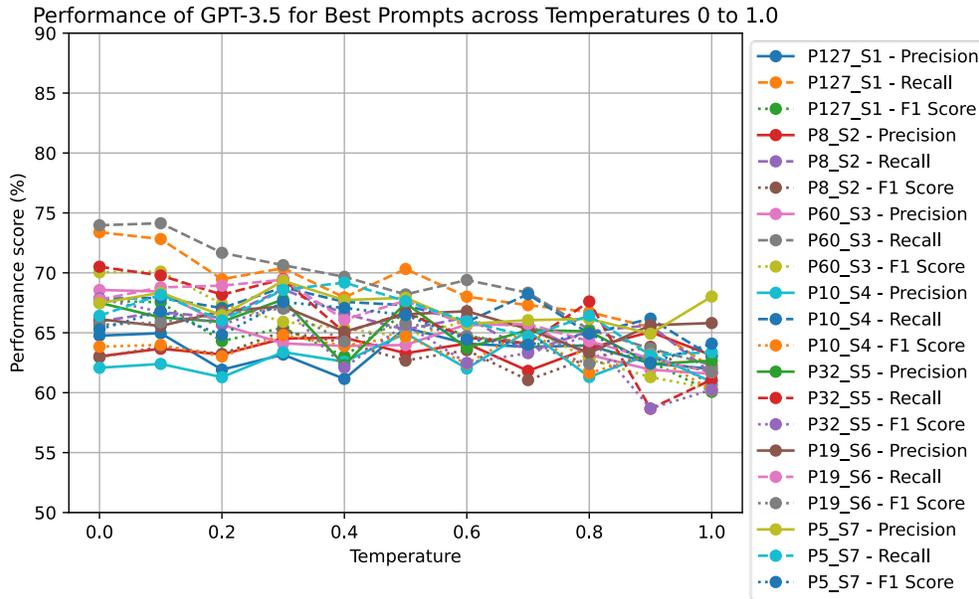

**Figure 4. Performance evaluation of temperature parameter of gpt-3.5-0613 for best prompts from each section.** (PX_SY means Prompt number X from Section number Y)

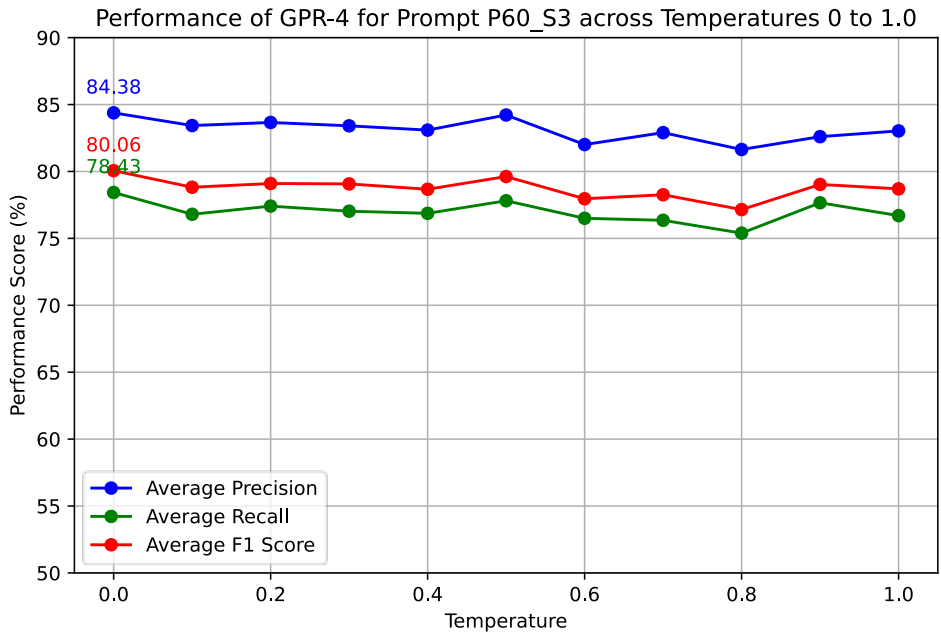

**Figure 5. Performance evaluation of temperature parameter of gpt-4-0613 for Prompt60 of Section 3.**

## PPI identification performance

The experimental analysis centers on the comparative performance of autoregressive language models, specifically GPT-3.5 and GPT-4 variants, against masked language models like BioBERT, PubMedBERT, and SciBERT in a PPI prediction task.

**Table 6** presents a comparative analysis of performance on the LLL dataset. GPT-4, especially when enhanced with a Protein dictionary, exhibits remarkable strengths. It stands out with its precision rate of 88.37%, which notably exceeds that of the BERT-based models. This high precision indicates GPT-4's ability to identify relevant information accurately, a crucial aspect in data processing and analysis. While its recall and F1 score are slightly lower than the BioBERT model, trailing by a mere 6.81% and 0.35%, respectively, this small margin highlights GPT-4's balanced proficiency in both precision and f1 score. Furthermore, GPT-4's recall rate, though lower than PubMedBERT's, is still impressive at 85.14%.

**Table 6: Evaluation result of PPI identification task on the LLL dataset for BERT and GPT-based models.**

| Type | Model | Precision | Recall | F1-Score |
|---|---|---|---|---|
| *Autoregressive Language Models* | GPT-3.5 with base prompt | 68.57% | 73.97% | 70.05% |
| | GPT-3.5 with protein dictionary | 79.06% | 75.95% | 76.72% |
| | GPT-3.5 with normalized protein dictionary | 74.20% | 70.55% | 71.39% |
| | GPT-3.5 with PROTEIN Masking | 66.90% | 47.52% | 52.68% |
| | GPT-3.5 with PROTEIN Masking – No Repeated Sentence in the same fold | 52.28% | 44.85% | 42.25% |
| | GPT-3.5 with PROTEIN Masking – one sentence at a time | 45.16% | 9.53% | 12.63% |
| | GPT-4 with base prompt | 84.38% | 78.43% | 80.06% |
| | GPT-4 with protein dictionary | **88.37%** | **85.14%** | **86.49%** |
| | GPT-4 with normalized protein dictionary | 87.97% | 83.25% | 85.21% |
| | GPT-4 with PROTEIN Masking | 73.99% | 62.48% | 64.72% |
| | GPT-4 with PROTEIN Masking – No Repeated Sentence in the same fold | 62.58% | 64.07% | 60.74% |
| | GPT-4 with PROTEIN Masking – one sentence at a time | 63.68% | 71.71% | 66.29% |
| *Masked Language Models* | BioBERT | 82.82% | **91.95%** | **86.84%** |
| | PubMedBERT | **85.25%** | 87.35% | 85.42% |
| | SciBERT | 84.54% | 86.07% | 84.66% |

**Table 7** presents a comparative analysis of the performance of the models on IEPA dataset. For this dataset, again, BERT-based models outperform GPT-based models on recall and F1 metrics. Among all settings of GPT-based models, GPT3.5 with protein masking (one sentence at a time) has the highest precision (78.95%), and GPT-4 with protein masking (one sentence at a time) has the highest recall (79.41%) and F1 score (71.54%). The use of domain-specific dictionaries and normalization improves the performance of the models, suggesting that incorporating domain knowledge is beneficial. However, despite these improvements, it is the masked language models that demonstrate superior overall efficacy, with the BioBERT model's 83.21% recall and 78.81% F1 score and the PubMedBERT model's 77.70% precision.

**Table 7: Evaluation result of PPI identification task on the IEPA dataset for BERT and GPT-based models.**

| Type | Model | Precision | Recall | F1-Score |
|---|---|---|---|---|
| *Autoregressive Language Models* | GPT-3.5 with base prompt | 17.58% | 53.35% | 25.53% |
| | GPT-3.5 with protein dictionary | 20.69% | 65.92% | 31.25% |
| | GPT-3.5 with normalized protein dictionary | 20.45% | 64.7% | 30.79% |
| | GPT-3.5 with PROTEIN Masking | 61.55% | 46.28% | 50.62% |
| | GPT-3.5 with PROTEIN Masking – No Repeated Sentence in the same fold | 57.74% | 39.97% | 39.34% |
| | GPT-3.5 with PROTEIN Masking – one sentence at a time | **78.95%** | 8.69% | 15.32% |
| | GPT-4 with base prompt | 23.66% | 64.94% | 34.19% |
| | GPT-4 with protein dictionary | 35.36% | 78.78% | 48.37% |
| | GPT-4 with normalized protein dictionary | 30.95% | 77.70% | 43.80% |
| | GPT-4 with PROTEIN Masking | 58.61% | 72.71% | 64.00% |
| | GPT-4 with PROTEIN Masking – No Repeated Sentence in the same fold | 55.06% | 62.03% | 55.89% |
| | GPT-4 with PROTEIN Masking – one sentence at a time | 65.71% | **79.41%** | **71.54%** |
| *Masked Language Models* | BioBERT | 75.80% | **83.21%** | 78.81% |
| | PubMedBERT | **77.70%** | 81.05% | 78.49% |
| | SciBERT | 73.29% | 79.32% | 75.53% |

The comparative performance of the models for HPRD50 dataset is listed in **Table 8**. GPT-4 with protein masking (one sentence at a time) impressively achieves the best recall 95.22% for HPRD50 dataset. However, both GPT-3.5 and GPT-4 lag BERT-based models in terms of precision and F1 score. Among all the BERT-based models, PubMedBERT has attained the highest performance metrics with a precision of 78.81%, a recall of 82.71%, and an F1 score of 79.65% for HPRD50 dataset. Thus, PubMedBERT has the highest overall efficacy among all of the models on HPRD50. In **Table 6**, **Table 7,** and **Table 8**, the performance scores of the top-performing models are highlighted in bold.

**Table 8: Evaluation result of PPI identification task on the HPRD50 dataset for BERT and GPT-based models.**

| Type | Model | Precision | Recall | F1-Score |
|---|---|---|---|---|
| *Autoregressive Language Models* | GPT-3.5 with base prompt | 40.62% | 73.89% | 50.96% |
| | GPT-3.5 with protein dictionary | 42.99% | 74.80% | 52.90% |
| | GPT-3.5 with normalized protein dictionary | 40.06% | 72.05% | 50.48% |
| | GPT-3.5 with PROTEIN Masking | 48.13% | 76.21% | 56.33% |
| | GPT-3.5 with PROTEIN Masking – No Repeated Sentence in the same fold | 44.7% | 89.29% | 57.26% |
| | GPT-3.5 with PROTEIN Masking – one sentence at a time | **66.40%** | 44.72% | 49.09% |
| | GPT-4 with base prompt | 44.83% | 75.16% | 55.33% |
| | GPT-4 with protein dictionary | 45.90% | 78.90% | 57.24% |
| | GPT-4 with normalized protein dictionary | 47.76% | 80.64% | 58.96% |
| | GPT-4 with PROTEIN Masking | 55.58% | 56.24% | 54.62% |
| | GPT-4 with PROTEIN Masking – No Repeated Sentence in the same fold | 58.06% | 71.35% | 61.24% |
| | GPT-4 with PROTEIN Masking – one sentence at a time | 50.36% | **95.22%** | 65.00% |
| *Masked Language Models* | BioBERT | 75.79% | 77.63% | 74.95% |
| | PubMedBERT | **78.81%** | **82.71%** | **79.65%** |
| | SciBERT | 73.03% | 79.02% | 74.67% |

Our analysis revealed interesting trends in the performance of different variations of LLMs, on which we focused on autoregressive language models and masked language models in PPI prediction tasks. Specifically, BERT-based models generally showed superior performance in complex datasets like IEPA and HPRD50 compared to GPT models.

## Discussion

It is worth noting that although BERT-based models demonstrate impressive performance, they require fine-tuning with supervised learning, which takes considerable time and technical expertise. In comparison, zero-shot learning models such as GPT-3, GPT-3.5, and GPT-4 do not require such extensive fine-tuning, making them more accessible and practical for specific use cases. This is because GPT models are much larger in terms of model parameters and pre-trained on more datasets compared to BERT-based models. Therefore, despite being primarily designed for text generation, GPT-3.5 and GPT-4 have demonstrated remarkable ability in identifying PPIs from biomedical literature.

The major difference among our six prompt settings is that for half of them (Base prompt, base prompt with protein dictionary, and base prompt with normalized protein dictionary), we used the original sentences as the input. For the other half of the settings, we used Protein masked sentences, namely 10-fold input (identical folds used for BERT-based models), N-fold input (with no repeated sentence in the same fold), and one sentence at a time as input. We conducted a thorough comparison of GPT and BERT-based models with identical 10-fold settings to maintain consistency across all models, as well as explored other possible settings.

From the experimental results, we observed that BERT-based models, including BioBERT, PubMedBERT, and SciBERT, tend to show less variability in performance across the different datasets compared to GPT-based models. This could be due to the variation of BERT-based models we have used, actually have the same model architecture. Another potential reason could be that the BERT-based models have been pre-trained on domain-specific corpora, which might make it more robust to variations in the protein relation datasets.

GPT-based models have competitive performance in many configurations but seem to be more sensitive to the prompt structure and dictionary type, possibly due to GPT's broader language model training.

The performance also varies across datasets for all the models, which implies that each dataset has its own set of challenges. For example, the IEPA dataset seems to have longer protein names on average (**Table 4**), which might affect model performance if the model has limitations in processing longer sequences. This may be the potential reason why IEPA and HPRD50 work better in the case of PROTEIN masked settings compared to original sentences. However, the use of variation in the same sentence for different PROTEIN1 PROTEIN2 location pairs seems to affect the performance. This could be indicative of the models' sensitivity to the sentences given in the same input despite explicitly informing it to consider each sentence separately. To address this, the approach of processing one sentence at a time was investigated, revealing that GPT-4 notably excels with more complex datasets like HPRD50 and IEPA, suggesting its superior handling of intricate dataset characteristics.

This study, although offering interesting insights, has significant constraints. Initially, our investigation is limited exclusively to GPT models, specifically versions 3.5 and 4, as well as a restricted assortment of three BERT models, namely BioBERT, PubMedBERT, and SciBERT. The limitation in the range of LLMs available may have influenced the scope of our discoveries. Furthermore, the datasets used were not large enough. The generalizability of our results to more diverse datasets may be compromised due to the limited breadth of our conclusions, which is influenced by the volume and variety of the data. Furthermore, the use of LLMs such as GPT comes with inherent challenges, notably the high computational and financial costs associated with their deployment and maintenance. This factor is particularly relevant when considering the practical implementation of these models in real-life situations.

However, further development of language model-based methods is needed to address sensitive functions in these areas. Nevertheless, improving GPT-4 with biomedical corpora like PubMed and PMC for PPI identification is warranted. With additional information, such as a dictionary, GPT has shown decent performance and can demonstrate substantially improved performance comparable to BERT-based models, indicating the potential use of GPT for these NLP tasks. Further research is needed to explore and enhance the capabilities of GPT-based models in the biomedical domain.

## Future Direction

Our future goal is to address the identified limitations by expanding our research to include a wider variety of models and larger, more diverse datasets. Additionally, we aim to explore strategies for the cost-effective implementation of LLMs in practical settings, ensuring that their application is not only theoretically sound but also financially viable. Another key direction for our future research involves investigating the role of ontology in enhancing PPI literature mining. This exploration will focus on understanding how ontological frameworks can support and improve the efficiency and accuracy of extracting relevant information from biomedical texts. For example, we have previously developed the Interaction Network Ontology (INO) and have applied the INO for the mining of gene-gene or protein-protein interactions [44-47]. Our study showed that the INO ontology-based literature mining enhanced the mining of the gene-gene or protein-protein interactions. Each ontology term is associated with a list of keywords supporting enhanced literature mining. Meanwhile, ontology also provides semantic relations and hierarchical structure among different interactions, which provides a basis for further interpretation and analysis of the mined interactions. While the currently reported study did not apply ontology, we plan to investigate how the ontology can be used together with existing literature mining tools to enhance our mining performance further.

## Acknowledgments

The study was supported by the US National Institute of Allergy and Infectious Disease (U24AI171008 to Y.H. and J.H.). GEBIP Award of the Turkish Academy of Sciences (to A.Ö.) is gratefully acknowledged.

# Supplementary Figures

**Supplementary Figure 1. PPI mentioned in a biomedical text.**

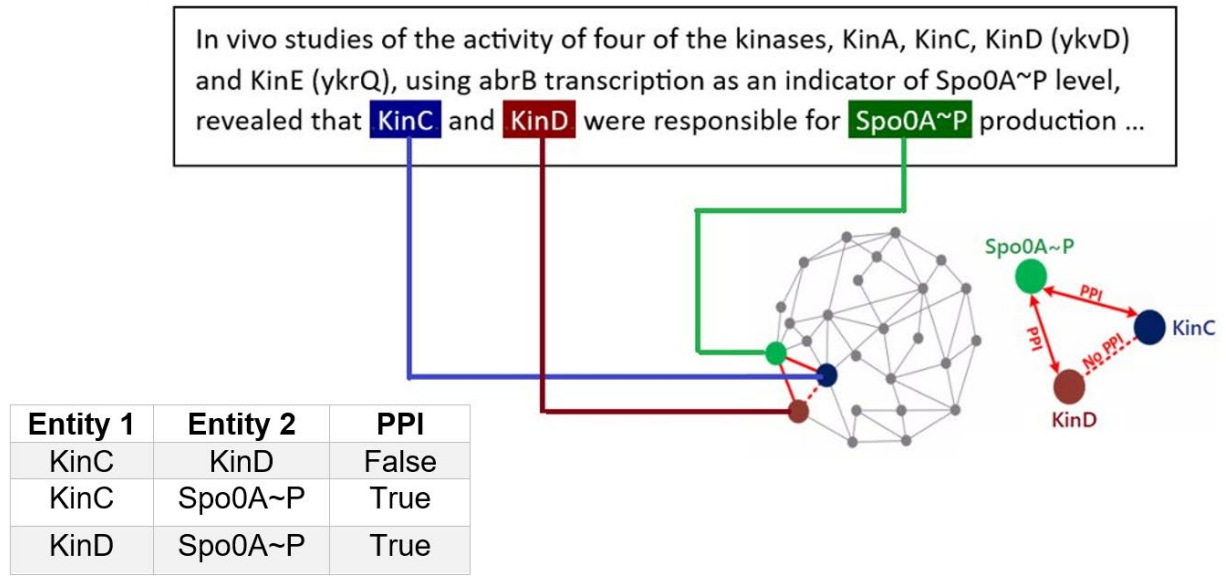

**Supplementary Figure 2. Overview of methodology.**

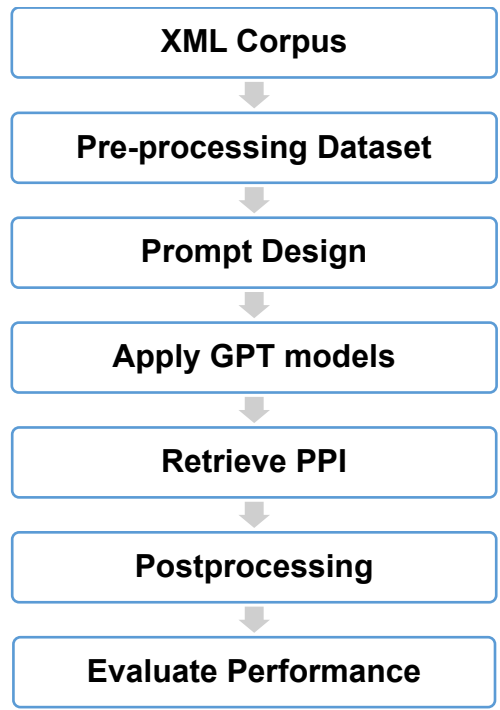

## Supplementary Figure 3. Prompt Engineering for Base Prompt- Section 1

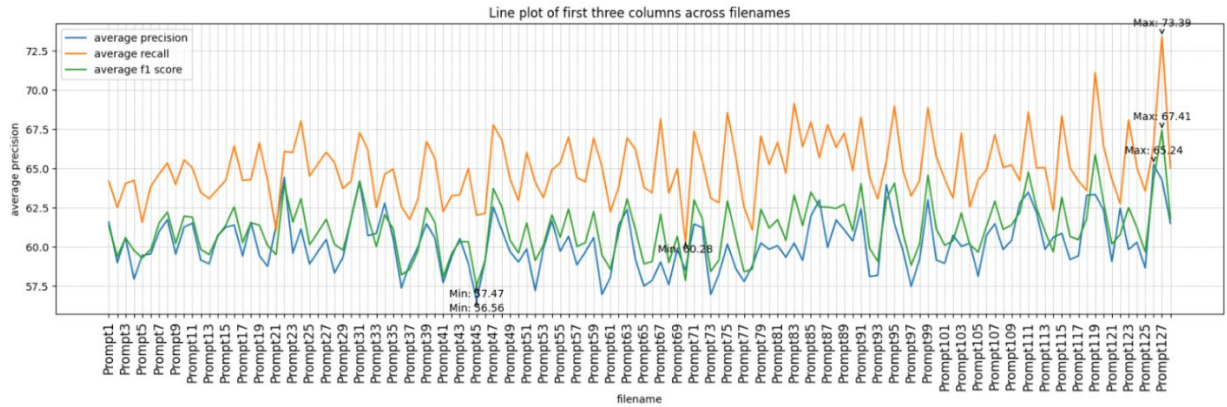

## Supplementary Figure 4. Prompt Engineering for Base Prompt- Section 2

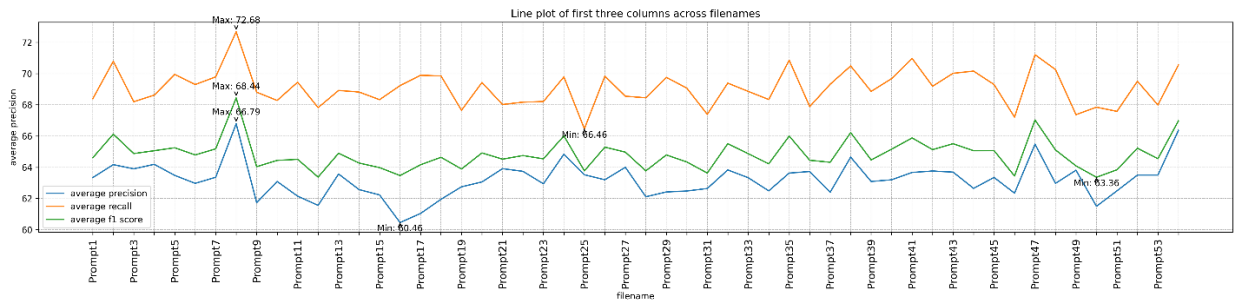

## Supplementary Figure 5. Prompt Engineering for Base Prompt- Section 3

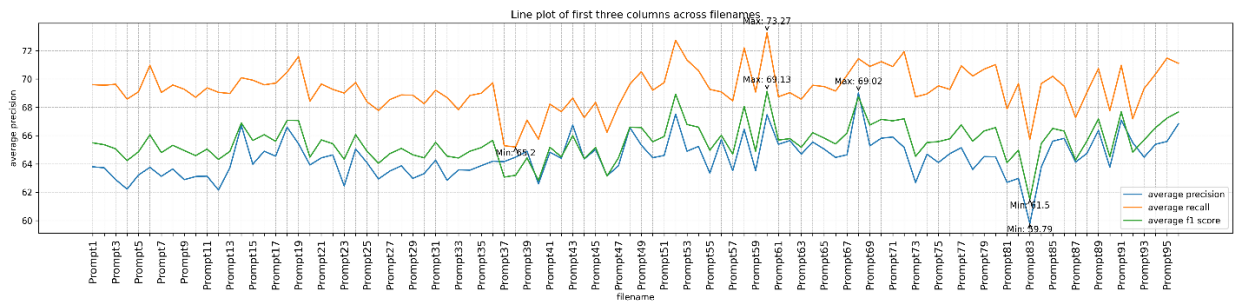

## Supplementary Figure 6. Prompt Engineering for Base Prompt- Section 4

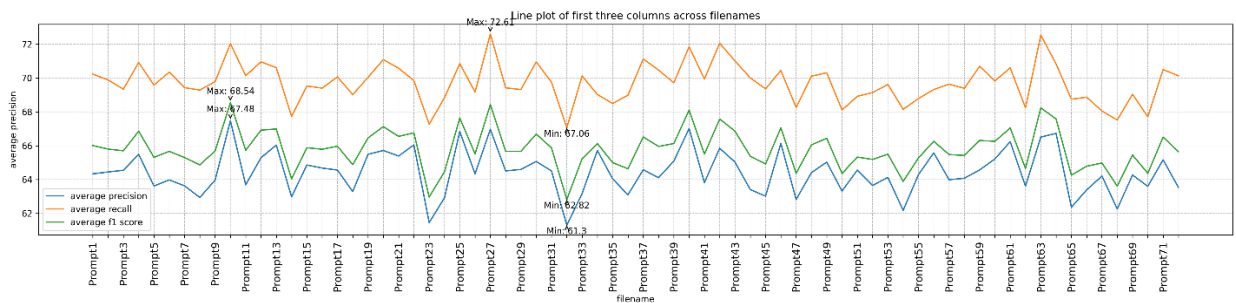

## Supplementary Figure 7. Prompt Engineering for Base Prompt- Section 5

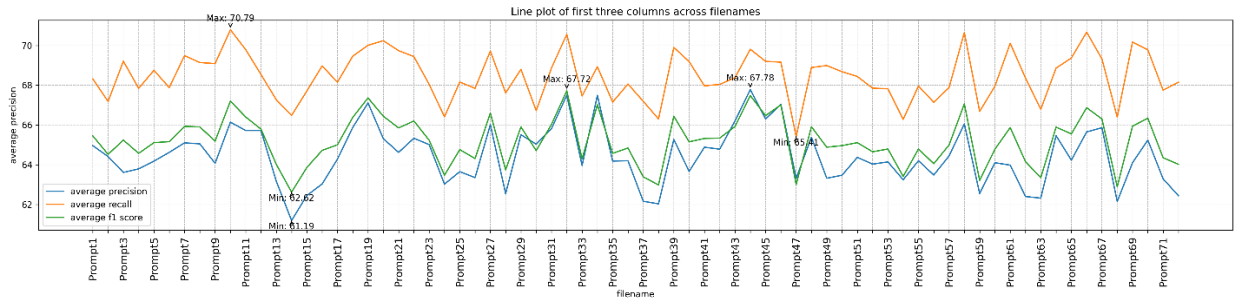

## Supplementary Figure 8. Prompt Engineering for Base Prompt- Section 6

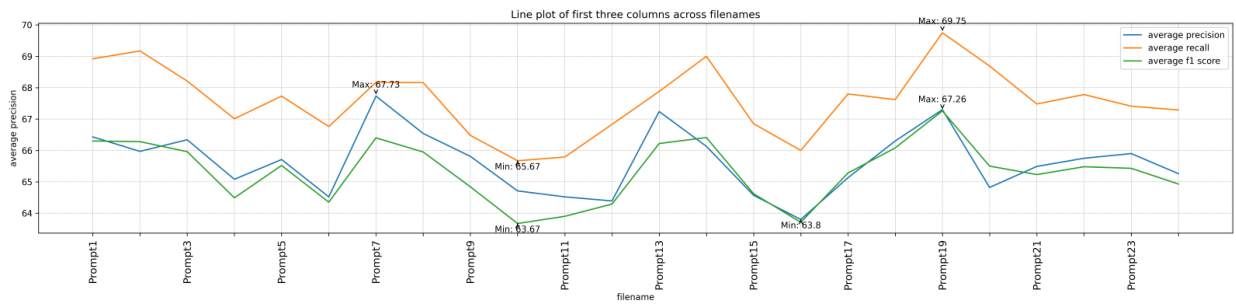

## Supplementary Figure 9. Prompt Engineering for Base Prompt- Section 7

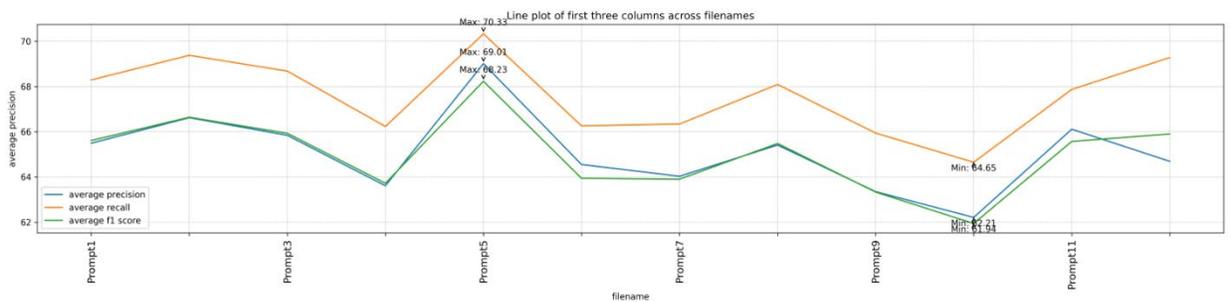

**Supplementary Figure 10. Prompt Engineering for PROTEIN masked Prompt**

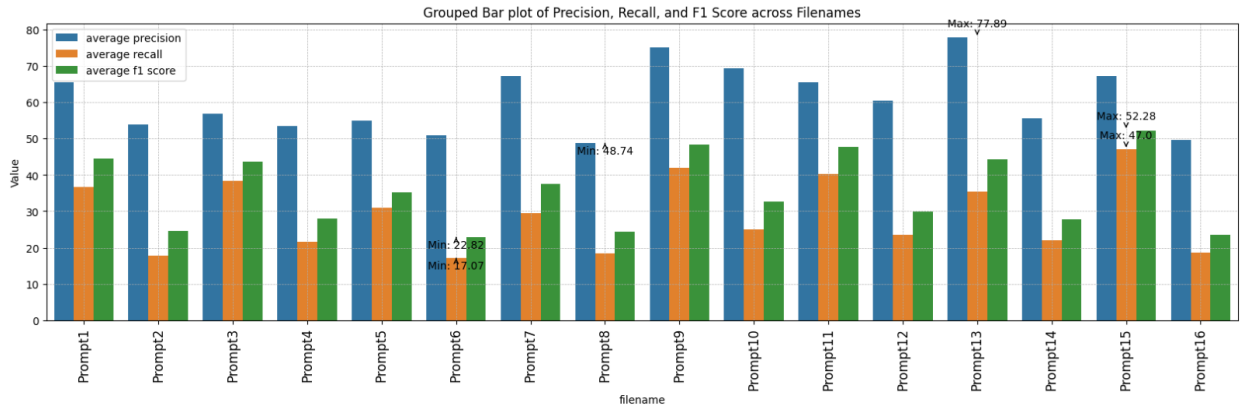

**Supplementary Figure 11. Evaluation result of PPI identification on LLL dataset for BERT and GPT Based Models**

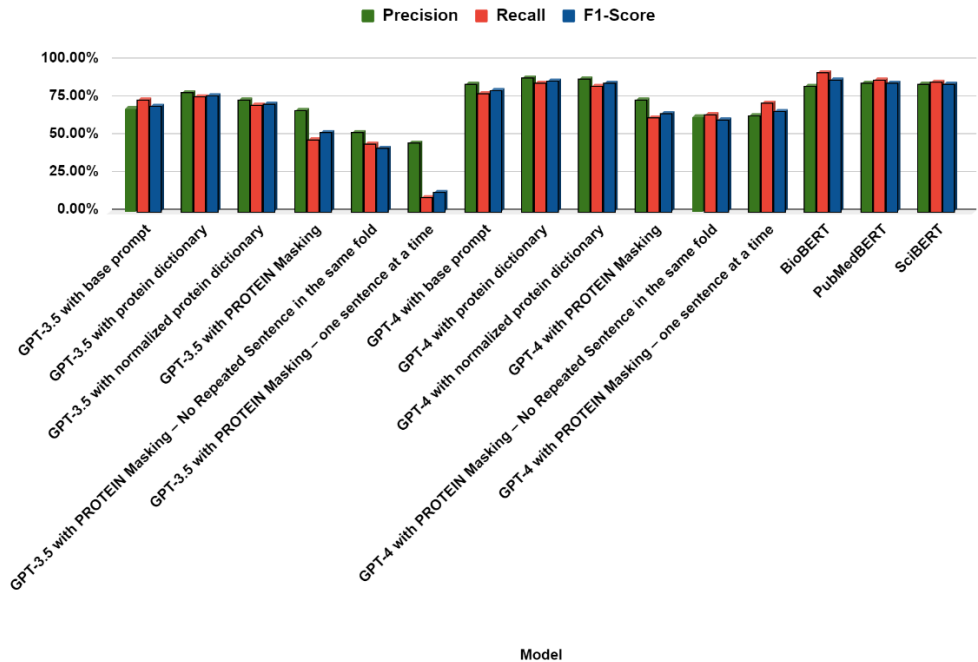

## Supplementary Figure 12. Evaluation result of PPI identification on HPRD50 dataset for BERT and GPT Based Models

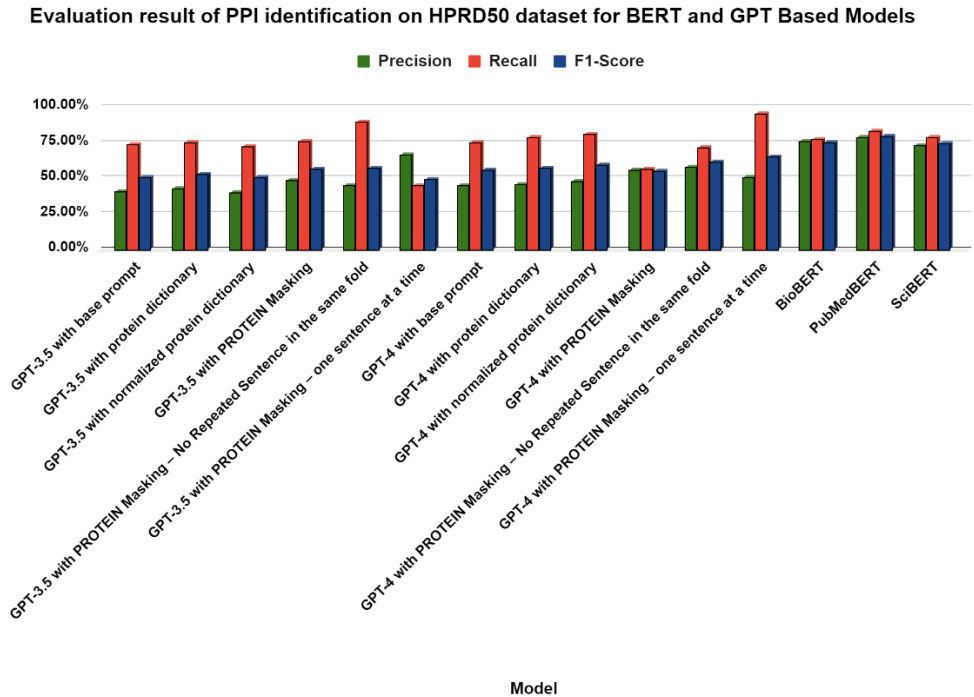

## Supplementary Figure 13. Evaluation result of PPI identification on IEPA dataset for BERT and GPT Based Models

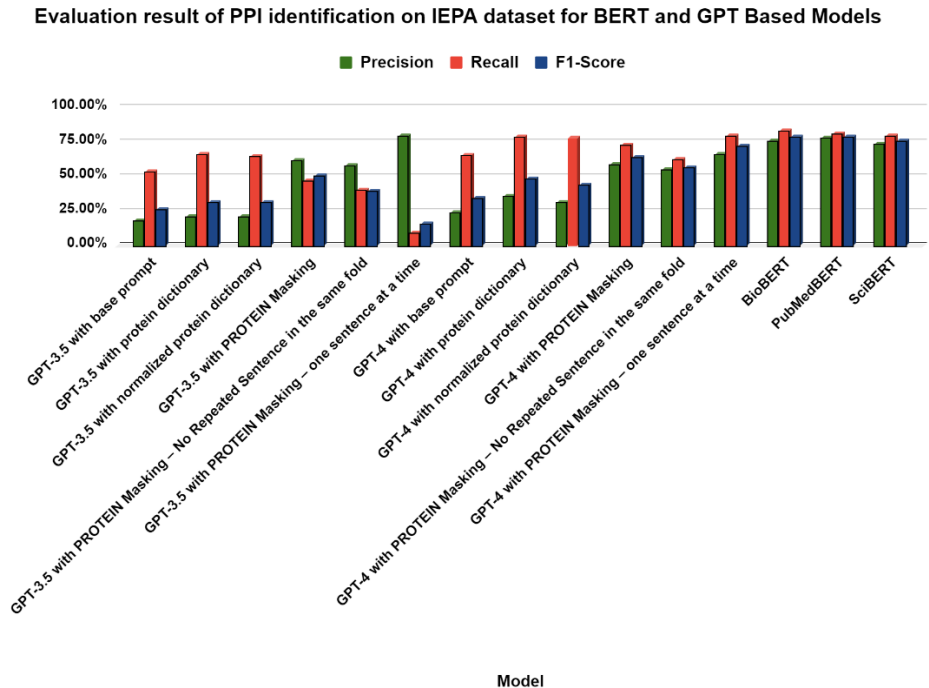

# Supplementary Tables

**Supplementary Table 1. Specifications of GPT models highlighting their corresponding training data, release year, structural architecture, parameter count, and context window capacity for input and output.**

| Model Name | Training data | Year released | Architecture | Number of Parameters | Context Window |
|---|---|---|---|---|---|
| GPT-1 | Common Crawl, BookCorpus | 2018 | Decoder architecture of transformer with 12 layers | 117 million | 1024 |
| GPT-2 | - Common Crawl, BookCorpus, WebText | 2019 | Decoder architecture of transformer with 48 layers | 1.5 billion | 2048 |
| GPT-3 | Common Crawl, BookCorpus, Wikipedia, Books, Articles, and more (Up to Oct. 2019) | 2020 | Same model and architecture as GPT-2 with 96 layers. Variations include Davinci[#], Babbage, Curie, and Ada. | 175 billion | 2,049 tokens |
| GPT-3.5 | Up to Sep. 2021 | 2022 | A combination of three models: code-davinci-002, text-davinci-002, and text-davinci-003. | 1.3 billion, 6 billion, and 175 billion | 4096 tokens (Default) |
| GPT-4 | Dataset is not not officially disclosed (Up to Sep. 2021) | 2023 | Fine-tuned using reinforcement learning from human feedback. | Supposedly 100 trillion | 8,192 tokens (Default) |

[#]**Used in the current study.**

**Supplementary Table 2. Prompt Engineering for Base Prompt- Best Prompts from each section.**

| Prompt | Prompt |
|---|---|
| **P127_S1** | Consider each sentence separately and infer every pair of Protein-Protein Interactions from the provided sentences.<br>For this task, 'Proteins' and 'Genes' are synonymous.<br>If a sentence contains multiple PPI pairs, list each pair on a distinct row.<br>Please, format your results in CSV (comma-separated values) format with the following four columns: 'Sentence ID', 'Protein 1', 'Protein 2', and 'Interaction Type'. Ensure that no columns are left blank. |

| | | |
|---|---|---|
| | | Output Column Specifications:
'Sentence ID': The unique identifier for each sentence.
'Protein 1' and 'Protein 2': The entities in the sentence, representing the proteins or genes.
'Interaction Type': The type of interaction identified between the protein entities (e.g., 'binds to', 'inhibits').
If all sentences have been processed successfully, the last row should only contain the word 'Done'.
Each input line contains a 'Sentence ID' and corresponding 'Sentence' that is needed to be analyzed for finding PPI.
Here are the sentences that you need to process: |
| | **P8_S2** | Consider each sentence separately and infer every pair of Protein-Protein Interactions from the provided sentences.
For this task, consider Proteins and Genes as interchangeable terms.
If a sentence contains multiple PPI pairs, list each pair on a distinct row.
Please, format your results in CSV (comma-separated values) format with the following four columns: 'Sentence ID', 'Protein 1', 'Protein 2', and 'Interaction Type'. Ensure that no columns are left blank.
Output Column Specifications:
'Sentence ID': The unique identifier for each sentence.
'Protein 1' and 'Protein 2': The entities in the sentence, representing the proteins or genes.
'Interaction Type': The type of interaction identified between the protein entities (e.g., 'binds to', 'inhibits').
If all sentences have been processed successfully, the last row should only contain the word 'Done'.
Each input line contains a 'Sentence ID' and corresponding 'Sentence' that is needed to be analyzed for finding PPI.
Here are the sentences that you need to process: |
| | **P60_S3** | Consider each sentence separately and infer every pair of Protein-Protein Interactions from the provided sentences.
For this task, consider Proteins and Genes as interchangeable terms.
Provide each pair in a separate row whenever a sentence contains multiple Protein-Protein interaction pairs.
Please, format your results in CSV (comma-separated values) format with the following four columns: 'Sentence ID', 'Protein 1', 'Protein 2', and 'Interaction Type'. Ensure that no columns are left blank.
Output Column Specifications:
'Sentence ID': The unique identifier for each sentence.
'Protein 1' and 'Protein 2': The entities in the sentence, representing the proteins or genes.
'Interaction Type': The type of interaction identified between the protein entities (e.g., 'binds to', 'inhibits').
If all sentences have been processed successfully, the last row should only contain the word 'Done'.
Each input line contains a 'Sentence ID' and corresponding 'Sentence' that is needed to be analyzed for finding PPI.
Here are the sentences that you need to process: |

| P10_S4 | Consider each sentence separately and infer every pair of Protein-Protein Interactions from the provided sentences. |
| --- | --- |
| | For this task, consider Proteins and Genes as interchangeable terms. |
| | Provide each pair in a separate row whenever a sentence contains multiple Protein-Protein interaction pairs. |
| | Please format the output in CSV with the following four columns: 'Sentence ID', 'Protein 1', 'Protein 2', and 'Interaction Type'. Ensure that no columns are left blank. |
| | Output Column Specifications: |
| | 'Sentence ID': The unique identifier for each sentence. |
| | 'Protein 1' and 'Protein 2': The entities in the sentence, representing the proteins or genes. |
| | 'Interaction Type': The type of interaction identified between the protein entities (e.g., 'binds to', 'inhibits'). |
| | If all sentences have been processed successfully, the last row should only contain the word 'Done'. |
| | Each input line contains a 'Sentence ID' and corresponding 'Sentence' that is needed to be analyzed for finding PPI. |
| | Here are the sentences that you need to process: |
| **P32_S5** | Consider each sentence separately and infer every pair of Protein-Protein Interactions from the provided sentences. |
| | For this task, consider Proteins and Genes as interchangeable terms. |
| | Provide each pair in a separate row whenever a sentence contains multiple Protein-Protein interaction pairs. |
| | Please format the output in CSV with the following four columns: 'Sentence ID', 'Protein 1', 'Protein 2', and 'Interaction Type'. Ensure that no columns are left blank. |
| | Output Column Specifications: |
| | 'Sentence ID': The unique ID for each sentence. |
| | 'Protein 1' and 'Protein 2': The entity pairs in the sentence representing the proteins or genes with potential PPI. |
| | 'Interaction Type': The type of interaction identified between the protein pairs (e.g., 'binds to', 'inhibits'). |
| | If all sentences have been processed successfully, the last row should only contain the word 'Done'. |
| | Each input line contains a 'Sentence ID' and corresponding 'Sentence' that is needed to be analyzed for finding PPI. |
| | Here are the sentences that you need to process: |
| **P19_S6** | Consider each sentence separately and infer every pair of Protein-Protein Interactions from the provided sentences. |
| | For this task, consider Proteins and Genes as interchangeable terms. |
| | Provide each pair in a separate row whenever a sentence contains multiple Protein-Protein interaction pairs. |
| | Please format the output in CSV with the following four columns: 'Sentence ID', 'Protein 1', 'Protein 2', and 'Interaction Type'. Ensure that no columns are left blank. |
| | Output Column Specifications: |
| | The last output row should exclusively contain the word 'Done' to indicate that all the sentences have been processed successfully. |
| | 'Protein 1' and 'Protein 2': The entity pairs in the sentence representing the proteins or genes with potential PPI. |
| | 'Interaction Type': The type of interaction identified between the protein pairs (e.g., 'binds to', 'inhibits'). |
| | If all sentences have been processed successfully, the last row should only contain the word 'Done'. |
| | Each input line contains a 'Sentence ID' and corresponding 'Sentence' that is needed to be analyzed for finding PPI. |
| | Here are the sentences that you need to process: |
| **P5_S7** | Consider each sentence separately and infer every pair of Protein-Protein Interactions from the provided sentences. |
| | For this task, consider Proteins and Genes as interchangeable terms. |
| | Provide each pair in a separate row whenever a sentence contains multiple Protein-Protein interaction pairs. |
| | Please format the output in CSV with the following four columns: 'Sentence ID', 'Protein 1', 'Protein 2', and 'Interaction Type'. Ensure that no columns are left blank. |
| | Output Column Specifications: |
| | The last output row should exclusively contain the word 'Done' to indicate that all the sentences have been processed successfully. |
| | 'Protein 1' and 'Protein 2': The entity pairs in the sentence representing the proteins or genes with potential PPI. |
| | 'Interaction Type': The type of interaction identified between the protein pairs (e.g., 'binds to', 'inhibits'). |
| | If all sentences have been processed successfully, the last row should only contain the word 'Done'. |
| | Each input line contains a 'Sentence ID' and its corresponding 'Sentence' for PPI analysis. |
| | Here are the sentences that you need to process: |

# Prompt name format: PPromptNo_SSectionNo

## Supplementary Table 3. Final Prompts for each Prompt Type.

| Prompt Type | Prompt |
|---|---|
| **Base (10 fold)** | Consider each sentence separately and infer every pair of Protein-Protein Interactions from the provided sentences. |
| | For this task, consider Proteins and Genes as interchangeable terms. |
| | Provide each pair in a separate row whenever a sentence contains multiple Protein-Protein interaction pairs. |
| | Please, format your results in CSV (comma-separated values) format with the following four columns: 'Sentence ID', 'Protein 1', 'Protein 2', and 'Interaction Type'. Ensure that no columns are left blank. |
| | Output Column Specifications: |
| | 'Sentence ID': The unique identifier for each sentence. |
| | 'Protein 1' and 'Protein 2': The entities in the sentence, representing the proteins or genes. |
| | 'Interaction Type': The type of interaction identified between the protein entities (e.g., 'binds to', 'inhibits'). |
| | If all sentences have been processed successfully, the last row should only contain the word 'Done'. |

| | Each input line contains a 'Sentence ID' and corresponding 'Sentence' that is needed to be analyzed for finding PPI. |
| | Here are the sentences that you need to process: |
| **With Protein Dictionary (10 fold)** | Please, format your results in CSV (comma-separated values) format with the following four columns: 'Sentence ID', 'Protein 1', 'Protein 2', and 'Interaction Type'. Ensure that no columns are left blank. |
| | Output Column Specifications: |
| | 'Sentence ID': The unique identifier for each sentence. |
| | 'Protein 1' and 'Protein 2': The entities in the sentence, representing the proteins or genes. |
| | 'Interaction Type': The type of interaction identified between the protein entities (e.g., 'binds to', 'inhibits'). |
| | If all sentences have been processed successfully, the last row should only contain the word 'Done'. |
| | Each input line contains a 'Sentence ID' and corresponding 'Sentence' that is needed to be analyzed for finding PPI. |
| | Here are the protein names for your reference : [['KinC' 'KinD' 'sigma(A)' 'Spo0A' 'SigE' 'SigK' 'GerE' 'sigma(F)' 'sigma(G)' 'SpoIIE' 'FtsZ' 'sigma(H)' 'sigma(K)' 'gerE' 'EsigmaF' 'sigmaB' 'sigmaF' 'SpoIIAB' 'SpoIIAA' 'SigL' 'RocR' 'sigma(54)' 'E sigma E' 'YfhP' 'SpoIIAA-P' 'sigmaK' 'sigmaG' 'ComK' 'FlgM' 'sigma X' 'sigma B' 'sigma(B)' 'sigmaD' 'SpoIIID' 'sigmaW' 'PhoP~P' 'AraR' 'sigmaH' 'yvyD' 'ClpX' 'Spo0' 'RbsW' 'DnaK' 'sigmaE' 'sigma W' 'sigmaA' 'sigma(X)' 'CtsR' 'Spo0A~P' 'spoIIG' 'ydhD' 'ykuD' 'ykvP' 'ywhE' 'spo0A' 'spoVG' 'rsfA' 'cwlH' 'KatX' 'katX' 'rocG' 'yfhS' 'yfhQ' 'yfhR' 'sspE' 'yfhP' 'bmrUR' 'ydaP' 'ydaE' 'ydaG' 'yfkM' 'sigma F' 'cot' 'sigK' 'cotD' 'sspG' 'sspJ' 'hag' 'comF' 'flgM' 'ykzA' 'CsbB' 'nadE' 'YtxH' 'YvyD' 'bkd' 'degR' 'cotC' 'cotX' 'cotB' 'sigW' 'tagA' 'tagD' 'tuaA' 'araE' 'sigmaL' 'spo0H' 'sigma G' 'sigma 28' 'sigma 32' 'spoIVA' 'PBP4*' 'RacX' 'Ytel' 'YuaG' 'YknXYZ' 'YdjP' 'YfhM' 'phrC' 'sigE' 'ald' 'kdgR' 'sigX' 'ypuN' 'clpC' 'ftsY' 'gsiB' 'sigB' 'sspH' 'sspL' 'sspN' 'tlp']] |
| | Here are the sentences that you need to process: |
| **With Normalized Protein Dictionary (10 fold)** | Please, format your results in CSV (comma-separated values) format with the following four columns: 'Sentence ID', 'Protein 1', 'Protein 2', and 'Interaction Type'. Ensure that no columns are left blank. |
| | Output Column Specifications: |
| | 'Sentence ID': The unique identifier for each sentence. |
| | 'Protein 1' and 'Protein 2': The entities in the sentence, representing the proteins or genes. |
| | 'Interaction Type': The type of interaction identified between the protein entities (e.g., 'binds to', 'inhibits'). |
| | If all sentences have been processed successfully, the last row should only contain the word 'Done'. |
| | Each input line contains a 'Sentence ID' and corresponding 'Sentence' that is needed to be analyzed for finding PPI. |
| | Here are the normalized protein names for your reference : [['kinc' 'kind' 'sigmaa' 'spo0a' 'sige' 'sigk' 'gere' 'sigmaf' 'sigmag' 'spoiie' 'ftsz' 'sigmah' 'sigmak' 'esigmaf' 'sigmab' 'spoiiab' 'spoiiaa' 'sigl' 'rocr' 'sigma54' 'esigmae' 'yfhp' 'spoiiaa-p' 'comk' 'flgm' 'sigmax' 'sigmad' 'spoiiid' 'sigmaw' 'phop~p' 'arar' 'yvyd' 'clpx' 'spo0' 'rbsw' 'dnak' 'sigmae' 'ctsr' 'spo0a~p' 'spoiig' 'ydhd' 'ykud' 'ykvp' 'ywhe' 'spovg' 'rsfa' 'cwlh' 'katx' 'rocg' 'yfhs' 'yfhq' 'yfhr' 'sspe' 'bmrur' 'ydap' 'ydae' 'ydag' 'yfkm' 'cot' 'cotd' 'sspg' 'sspj' |
| | 'hag' 'comf' 'ykza' 'csbb' 'nade' 'ytxh' 'bkd' 'degr' 'cotc' 'cotx' 'cotb' 'sigw' 'taga' 'tagd' 'tuaa' 'arae' 'sigmal' 'spo0h' 'sigma28' 'sigma32' 'spoiva' 'pbp4*' 'racx' 'ytei' 'yuag' 'yknxyz' 'ydjp' 'yfhm' 'phrc' 'ald' 'kdgr' 'sigx' 'ypun' 'clpc' 'ftsy' 'gsib' 'sigb' 'ssph' 'sspl' 'sspn' 'tlp']]<br>Here are the sentences that you need to process: |
| **With PROTEIN masking (10-fold)** | Consider each sentence separately and infer Protein-Protein Interaction for the protein entity pairs PROTEIN1-PROTEIN2 from the provided sentences. Do not consider any other PROTEIN pairs in the sentence. In each sentence, original protein or gene names have been substituted with 'PROTEIN1', 'PROTEIN', and 'PROTEIN' placeholders. The placeholders used may represent a variety of proteins or genes, differing with each sentence.<br>Please, format your results in CSV (comma-separated values) with only two columns: 'Sentence ID' and 'PPI'. Do not include the original sentences or any explanation in the output.<br>Output Column Specifications:<br>'Sentence ID': The unique identifier for each sentence.<br>'PPI': Record your findings as 'TRUE' if there is a demonstrable interaction between PROTEIN1 and PROTEIN2, and 'FALSE' if there is none.<br>If all sentences have been processed successfully, the last row should only contain the word 'Done'.<br>Each input line contains a 'Sentence ID' and corresponding 'Sentence' that is needed to be analyzed for finding PPI.<br>Here are the sentences that you need to process: |
| **With PROTEIN masking (Nfold)** | Same as With PROTEIN masking (10-fold) |
| **With PROTEIN masking (One sentence at a time)** | Infer Protein-Protein Interaction for the protein entity pairs PROTEIN1-PROTEIN2 from the provided sentence. Do not consider any other PROTEIN pairs in the sentence. Original protein or gene names have been substituted with 'PROTEIN1', 'PROTEIN', and 'PROTEIN' placeholders.<br>Output Specification: Record your findings as 'TRUE' if there is a demonstrable interaction between PROTEIN1 and PROTEIN2, and 'FALSE' if there is none.<br>Here is the sentence that you need to process: |

**# 10 fold - Sentences in each fold here are the same as BERT, Nfold – The same sentence with different positional PROTEIN masking is not present in the same fold.**

**Supplementary Table 4. PPI identification macro scores of the BERT-based models on LLL**

| Model Name | Macro Precision | Macro Recall | Macro F1 Score |
|---|---|---|---|
| BioBERT | 86.80 | 85.92 | 85.66 |
| PubMedBERT | 86.44 | 85.08 | 84.79 |
| SciBERT | 86.16 | 82.95 | 83.52 |

**Supplementary Table 5. PPI identification macro scores of the BERT-based models on IEPA**

| Model Name | Macro Precision | Macro Recall | Macro F1 Score |
|---|---|---|---|
| BioBERT | 81.55 | 83.21 | 78.81 |
| PubMedBERT | 82.20 | 81.56 | 81.15 |
| SciBERT | 79.08 | 79.55 | 78.72 |

**Supplementary Table 6. PPI identification macro scores of the BERT-based models on HPRD50**

| Model Name | Macro Precision | Macro Recall | Macro F1 Score |
|---|---|---|---|
| BioBERT | 75.85 | 72.39 | 71.72 |
| PubMedBERT | 82.60 | 80.82 | 80.78 |
| SciBERT_scivocab_cased | 71.48 | 69.88 | 68.74 |